\documentclass[sigconf]{acmart}

\usepackage{caption} 
\usepackage{tikz}
\usepackage{amsmath}
\usepackage{balance}
\usepackage{bbm}
\usepackage{xspace}
\usepackage{float}
\usepackage{graphicx}

\usepackage{tikz}
\usetikzlibrary{calc}

\usepackage{filecontents}
\usepackage{subfigure}
\usepackage{listings}
\usepackage{color}
\usepackage{multirow}
\usepackage{stfloats}
\usepackage{placeins}

\renewcommand\footnotetextcopyrightpermission[1]{} 

\definecolor{mGreen}{rgb}{0,0.6,0}
\definecolor{mGray}{rgb}{0.5,0.5,0.5}
\definecolor{mPurple}{rgb}{0.58,0,0.82}
\definecolor{backgroundColour}{rgb}{0.95,0.95,0.92}

\newcommand{\up}{\vspace*{-0.5em}}

\lstdefinestyle{PythonStyle}{
	language=Python,
	backgroundcolor=\color{backgroundColour},
	commentstyle=\color{mGreen},
	keywordstyle=\color{magenta},
	numberstyle=\tiny\color{mGray},
	stringstyle=\color{mPurple},
	basicstyle=\ttfamily\linespread{0.9}\small,
	breakatwhitespace=false,
	breaklines=true,
	captionpos=b,
	keepspaces=true,
	numbers=left,
	numbersep=5pt,
	showspaces=false,
	showstringspaces=false,
	showtabs=false,
	tabsize=2,
    frame=lrtb
}
\newif\ifdraft
\drafttrue
\draftfalse

\ifdraft
  \newcommand{\zhao}[1]{{\textcolor{red}    { ***Zhao:     #1 }}}
  
  \newcommand{\outline}[1]{{\textcolor{blue} { ***Outline:  #1 }}}
  \newcommand{\shivaram}[1]{{\textcolor{cyan}      { ***Shivaram:      #1 }}}
  \newcommand{\ian}[1]{{\textcolor{green}    { ***Ian:      #1 }}}
  \newcommand{\kyle}[1]{{\textcolor{orange}   { ***Kyle:      #1 }}}
  \newcommand{\note}[1]{ {\textcolor{red}    { \bf          #1 }}}
  \newcommand{\greg}[1]{{\textcolor{magenta} { ***Greg:     #1 }}}
  \newcommand{\revision}[1]{{\textcolor{blue}{#1}}}
\else
  \newcommand{\zhao}[1]{}
  
  \newcommand{\outline}[1]{}
  \newcommand{\shivaram}[1]{}
  \newcommand{\ian}[1]{}
  \newcommand{\kyle}[1]{}
  \newcommand{\note}[1]{}
  \newcommand{\greg}[1]{}
  \newcommand{\revision}[1]{{#1}}
\fi

\newcommand{\name}{KAISA\xspace}

\copyrightyear{2021}
\acmYear{2021}
\setcopyright{acmlicensed}
\acmConference[SC '21]{The International Conference for High Performance Computing, Networking, Storage and Analysis}{November 14--19, 2021}{St. Louis, MO, USA}
\acmBooktitle{The International Conference for High Performance Computing, Networking, Storage and Analysis (SC '21), November 14--19, 2021, St. Louis, MO, USA}
\acmPrice{15.00}
\acmDOI{10.1145/3458817.3476152}
\acmISBN{978-1-4503-8442-1/21/11}

\begin{document}


\title{\name{}: An Adaptive Second-Order Optimizer Framework for Deep Neural Networks}

\author{J. Gregory Pauloski}
\affiliation{
    \institution{University of Chicago}
    \city{}
    \country{}
}
\author{Qi Huang}
\affiliation{
    \institution{University of Texas at Austin}
    \city{}
    \country{}
}
\author{Lei Huang}
\affiliation{
    \institution{Texas Advanced Computing Center}
    \city{}
    \country{}
}
\author{Shivaram Venkataraman}
\affiliation{
    \institution{University of Wisconsin, Madison}
    \city{}
    \country{}
}
\author{Kyle Chard}
\affiliation{
    \institution{University of Chicago}
    \institution{Argonne National Laboratory}
    \city{}
    \country{}
}
\author{Ian Foster}
\affiliation{
    \institution{University of Chicago}
    \institution{Argonne National Laboratory}
    \city{}
    \country{}
}
\author{Zhao Zhang}
\affiliation{
    \institution{Texas Advanced Computing Center}
    \city{}
    \country{}
}

\begin{abstract}

Kronecker-factored Approximate Curvature (K-FAC) has recently been shown to converge faster in deep neural network (DNN) training than stochastic gradient descent (SGD); however, K-FAC's larger memory footprint hinders its applicability to large models. 
We present \name{}, a {\bf K}-FAC-enabled, {\bf A}daptable, {\bf I}mproved, and {\bf S}c{\bf A}lable second-order optimizer framework that adapts the memory footprint, communication, and computation given specific models and hardware to improve performance and increase scalability. 
We quantify the tradeoffs between memory and communication cost and evaluate \name{} on large models, including ResNet-50, Mask R-CNN, U-Net, and BERT, on up to 128 NVIDIA A100 GPUs. 
\revision{Compared to the original optimizers, \name{} converges 18.1–36.3\% faster across applications with the same global batch size.
Under a fixed memory budget, \name{} converges 32.5\% and 41.6\% faster in ResNet-50 and BERT-Large, respectively. 
\name{} can balance memory and communication to achieve scaling efficiency equal to or better than the baseline optimizers.}
\name{} is open source and available at \url{https://github.com/gpauloski/kfac_pytorch}.

\end{abstract}


\maketitle

\pagestyle{plain} 

\section{Introduction}
\label{sec:intro}

Deep neural networks (DNNs) have driven breakthroughs in many research domains, including image classification~\cite{he2016deep, huang2017densely}, object detection and segmentation~\cite{he2017mask}, machine translation~\cite{wu2016google}, and language modeling~\cite{devlin2018bert}.
DNNs are typically trained with stochastic gradient descent (SGD), or variants thereof.
As models and training datasets become larger, training must 
increasingly be performed in parallel on many CPUs, GPUs, or TPUs~\cite{you2019large, kumar2019scale}.
For example, the BERT~\cite{devlin2018bert} model with 330~million parameters would take weeks to months to train on a single GPU, and GPT-3~\cite{brown2020language} with 175~billion parameters cannot fit in the memory of any commercially available GPU.
While distributed training on many processors can reduce training time,
the need to communicate weight updates and other information among processors can limit scalability~\cite{narayanan2019pipedream}.

\begin{figure}[t]
\begin{center}
	\includegraphics[width=\columnwidth,trim=7mm 7mm 6.8mm 6mm,clip]{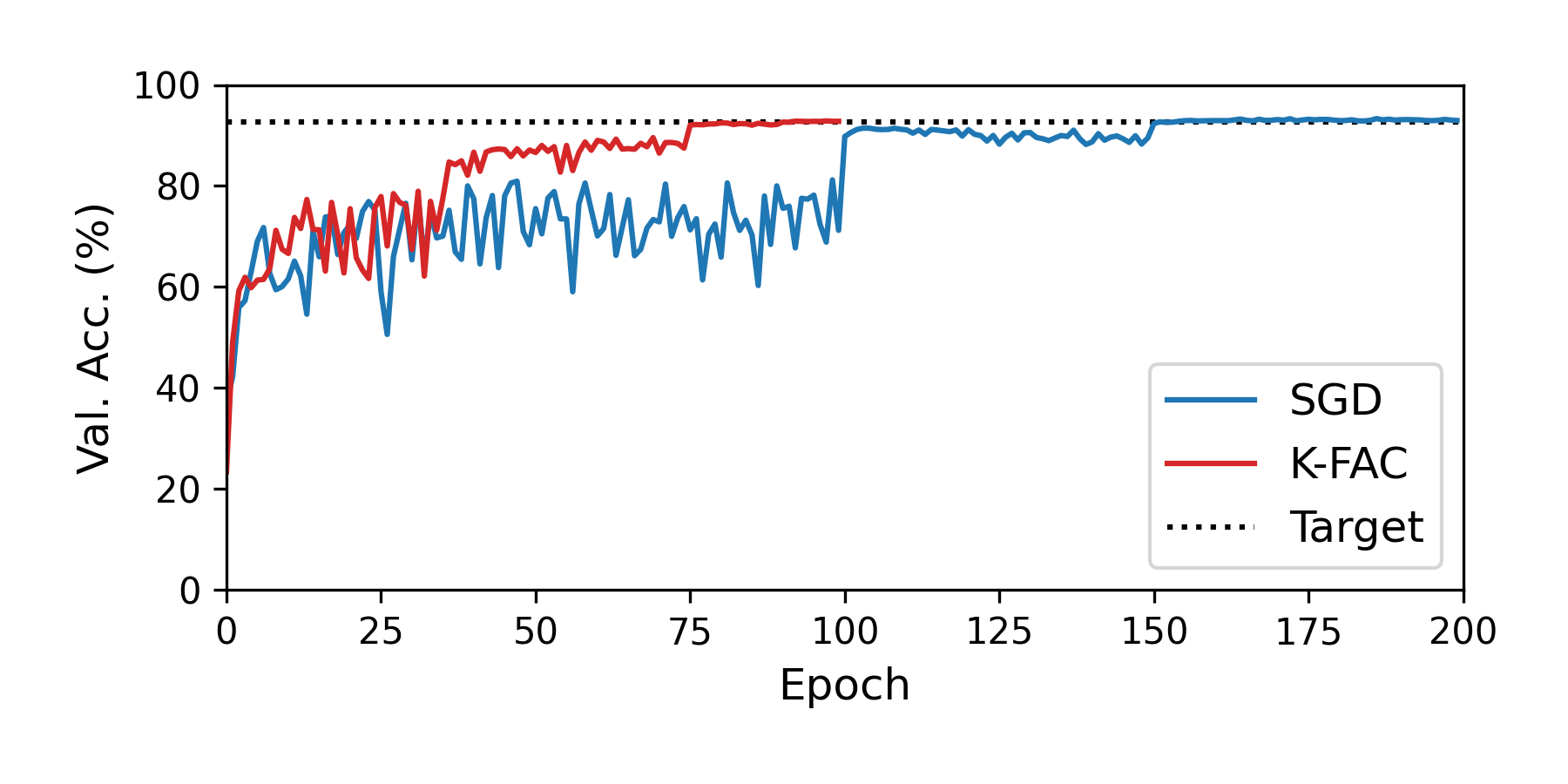}
	\caption{SGD vs. K-FAC training for ResNet-32 with the CIFAR-10 dataset. K-FAC reduces iterations needed for convergence.}
	\label{fig:cifar}
\end{center}
\end{figure}

Recent theoretical~\cite{martens2015optimizing, martens2016conv, martens2018kronecker} and empirical~\revision{\cite{Ba2017Distributed, Osawa_2019_CVPR, ueno2020kfac}} studies have shown
that Kronecker-factored Approximate Curvature (K-FAC) can accelerate training by enabling convergence with fewer iterations than SGD---for example, achieving baseline validation accuracy in 40\% fewer epochs than SGD on ResNet-32 with the CIFAR-10 dataset (see Figure~\ref{fig:cifar}).
Technically, researchers use K-FAC to approximate the inverse of the Fisher Information Matrix (FIM)---an approximation of the Hessian---and precondition the gradients before parameter updates~\cite{martens2015optimizing}.
The K-FAC approximation is 1) compute intensive, due to the required inverse or eigen decomposition calculations with $O(N^3)$ complexity ($N$ is the number of parameters);
2) memory intensive, as it stores per-layer activations and gradients, which may take $O(N^2)$ space compared to $O(N)$ for SGD; and
3) communication intensive for distributed training because the Kronecker factors and accompanying inverses or eigen decompositions must be communicated to all workers.
To overcome these overheads and achieve faster training times than SGD at scale, K-FAC updates are often decoupled from gradient preconditioning so the expensive computations are performed less frequently.

Existing K-FAC systems either cache all Kronecker factor eigen decompositions needed to precondition gradients locally~\cite{pauloski2020convolutional} or distribute gradient preconditioning across processors~\cite{Osawa_2019_CVPR}. 
Neither approach enables K-FAC DNN deployments that are both memory and communication efficient, thus inhibiting scalability.
The first approach avoids communication when local memory is sufficient.
The second approach reduces memory footprint by not caching eigen decompositions locally but increases communication.

These various considerations make it difficult to use K-FAC efficiently in practice given the specific requirements of the DNN model or hardware and require users to choose between optimizing memory or communication.
To address these concerns, we propose \name{}, a {\bf K}-FAC-enabled, {\bf A}daptable, {\bf I}mproved, and {\bf S}c{\bf A}lable second-order optimizer framework that can adapt execution to a given model size and memory limit.
\name{} allows tuning the ratio between communication and memory footprint to optimally apply K-FAC to distributed training by controlling the fraction of processes with local access to data.
This adaptation is made possible by grouping the processes, then distributing and communicating data within each group.
K-FAC distribution strategies described in other work are effectively special cases of \name{} in which either all processes~\cite{pauloski2020convolutional} or one process~\cite{Osawa_2019_CVPR} cache data.

We evaluate \name{} on a range of classification, segmentation, and language modeling applications, including ResNet-50~\cite{he2016deep}, Mask R-CNN~\cite{he2017mask}, U-Net~\cite{ronneberger2015u}, and BERT~\cite{devlin2018bert}, on clusters of 448 NVIDIA Tesla V100s and 192 Ampere A100 GPUs.
Our results show that:
1) with fixed global batch size, \name{} trains deep neural networks 18.1--36.3\% faster than \revision{the original optimizers} for these applications;
\revision{2) with a fixed memory budget, \name{} trains ResNet-50 and BERT-Large phase 2 in 32.5\% and 41.6\% less time compared to momemtum SGD and Fused LAMB, respectively;}
3) by varying the number of gradient workers, the K-FAC \revision{memory} overhead can be reduced by \revision{1.5--$2.9\times$};
4) for high communication applications such as ResNet-50, \revision{extra processor memory can be used to train 24.4\% faster}; and
5) using state-of-the-art NVIDIA A100 GPUs, \name{} converges in fewer iterations at all scales, and \name{}'s scaling performance is on-par with SGD even with \name{}'s  additional communication overhead.
Our code is open source with the MIT license and available at \url{https://github.com/gpauloski/kfac_pytorch}.

Our contributions in this paper are:
\begin{itemize}
  \item{} An adaptive second-order optimizer framework that trains faster than SGD and its variants, while preserving convergence.
  \item{} A quantitative study of the tradeoff between memory footprint and communication and its impact on training time in K-FAC design.
  \item{} The first large scale evaluation of K-FAC convergence and speedup relative to SGD for Mask R-CNN, U-Net, and BERT on up to 128 A100 GPUs.
\end{itemize}

The rest of the paper is as follows.
We present the mathematical background and distributed implementation of K-FAC in \S\ref{sec:back}.
We describe \name{}'s design in \S\ref{sec:design} and implementation in \S\ref{sec:impl}.
We present our experiments and results in \S\ref{sec:experiments}.
In \S\ref{sec:related}, we summarize existing DNN optimization frameworks and memory management techniques.
Finally, we conclude in \S\ref{sec:conc}. 

\section{Background}
\label{sec:back}

We first introduce K-FAC and its distributed implementation.
K-FAC is an efficient approximation of the Fisher Information Matrix (FIM), which has been shown to be equivalent to the Generalized Gauss-Newton (GGN) matrix in specific cases and can be viewed as an approximation of the Hessian~\cite{martens2015optimizing}.
In a standard SGD update step, the weights are updated using the gradients of the loss, as illustrated in Equation~\ref{eq:sgd}.
The K-FAC update step in Equation~\ref{eq:kfac} uses the FIM $F$ to precondition the gradients prior to update~\revision{\cite{pauloski2020convolutional}}.

$w^{(k)}$ is the weight at iteration $k$, ${\alpha}^{(k)}$ is the learning rate at iteration $k$, $n$ is the mini-batch size, $\nabla{L_i}(w^{(k)})$ is the gradient of the loss function $L_i$ for the $i^\text{th}$ example with regard to $w^{(k)}$, and $F^{-1}(w^{(k)})$ is the inverse of the FIM.
\begin{align}
\text{SGD:  } w^{(k+1)} &= w^{(k)} - \frac{{\alpha}^{(k)}}{n}\sum_{i=1}^{n}\nabla{L_i}(w^{(k)}) \label{eq:sgd}\\
\text{K-FAC: } w^{(k+1)} &= w^{(k)} - \frac{{\alpha}^{(k)}F^{-1}(w^{(k)})}{n}\sum_{i=1}^{n} \nabla{L_i}(w^{(k)}) \label{eq:kfac}
\end{align}

It has been shown empirically that training DNNs with the K-FAC second-order method enables convergence with fewer iterations than with SGD alone.
Theoretical understandings of the convergence rates of natural gradient methods, such as K-FAC, are an area of active research.
Previous work has shown that K-FAC~\cite{zhang2019fast} has linear convergence to the global minimum given a sufficiently over-parameterized model.
In strongly-convex problems, natural gradient methods have a quadratic convergence rate compared to the linear convergence of SGD~\cite{bottou2018optimization}.
The strongly-convex case provides some understanding of how K-FAC can improve convergence in non-convex cases.
Further, it has been shown that natural gradient methods enable larger learning rates improving the rate of convergence~\cite{zhang2019fast}.
K-FAC makes greater per-iteration progress in minimizing the objective function at the cost of more computationally expensive iterations.

\subsection{K-FAC Approximation}
K-FAC is based on the Kronecker product, a block matrix factorization that can reduce a large matrix inverse into two smaller matrix inverses.
K-FAC exploits the properties of the Kronecker product and the geometry of the FIM for DNNs to greatly reduce the complexity of computing the approximate FIM inverse.

\subsubsection{Kronecker Product}
The Kronecker product is written as $A \otimes B$ where $A$ has size $m\times n$ and $B$ has size $p \times q$.
The resulting matrix has shape $mp\times nq$.
\begin{align}
A \otimes B=
\begin{bmatrix}
a_{11}B  & \dots  & a_{1n}B \\
\vdots   & \ddots & \vdots  \\
a_{m1}B  & \dots  & a_{mn}B
\end{bmatrix}
\end{align}

The Kronecker product has two convenient properties: 
\begin{align}
(A \otimes B)^{-1} &= A^{-1} \otimes B^{-1} \label{eq:prop1}\\
(A \otimes B)\vec{c} &= B^\top \vec{c} A \label{eq:prop2}.
\end{align}

\subsubsection{K-FAC Approximation}

\begin{figure}[]
\begin{center}
    \includegraphics[width=0.9\columnwidth]{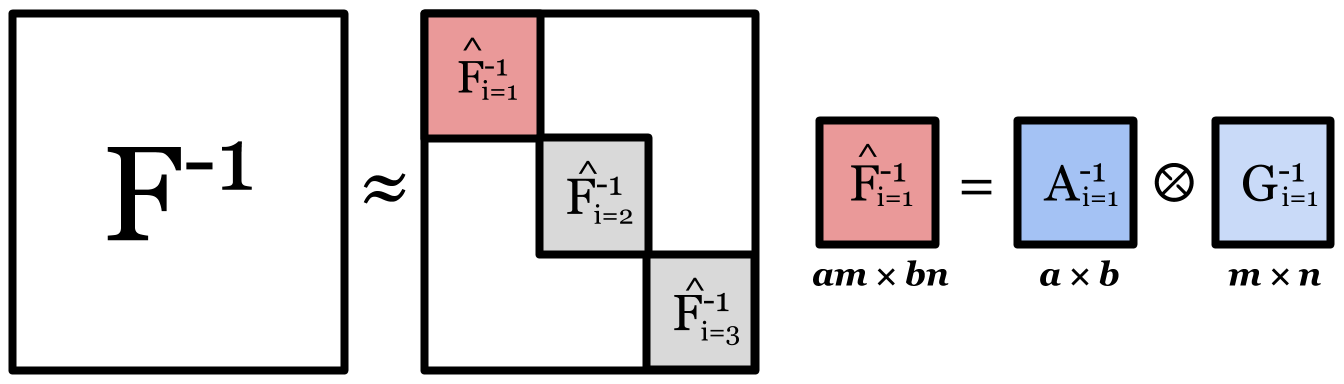}
    \caption{The K-FAC approximation of the Fisher information matrix. $\otimes$ is the Kronecker product~\revision{\cite{pauloski2020convolutional}}.}
    \label{fig:kfac}
\end{center}
\end{figure}

The FIM for a deep neural network is a block matrix where each block maps to layers in the model.
The block corresponding to the $i$ and $j^{\text{th}}$ layers is approximated as
\begin{equation}
F_{i,j}\approx a_{i-1}a_{j-1}^\top\otimes g_{i}g_{j}^\top.
\end{equation}
Here, $a_{i-1}$ and $g_i$ are the activation of the $i-1^\text{th}$ layer and the gradients of the $i^\text{th}$ layer in the model, respectively~\revision{\cite{martens2015optimizing, pauloski2020convolutional}}.\footnote{Formally, $F$ represents an expected value, and the expectation of a Kronecker product is not equivalent to the Kronecker product of the expected factors.
However, this approximation still reasonably represents the structure of the FIM~\cite{martens2015optimizing}.}

A fundamental assumption of K-FAC is the independence between layers~\revision{\cite{martens2015optimizing, pauloski2020convolutional}}.\footnote{This assumption is sufficient to produce an effective approximation for $F$ and necessary to produce a tractable algorithm.}
Using this assumption, K-FAC approximates the FIM as a diagonal block matrix $\hat{F}$.
\begin{equation}
\hat{F}=\texttt{diag}(\hat{F}_1,...,\hat{F}_i,...,\hat{F}_L)
\end{equation}

As shown in Figure \ref{fig:kfac}, the inverse of $F$ is a diagonal block matrix composed of the inverses of each diagonal block $\hat{F}_i$:
\begin{equation}
\hat{F}^{-1}=\texttt{diag}(\hat{F}_1^{-1},...,\hat{F}_i^{-1},...,\hat{F}_L^{-1})
\end{equation}
where
\begin{equation}
\hat{F}_{i}= a_{i-1}a_{i-1}^\top\otimes g_{i}g_{i}^\top=A_{i-1}\otimes G_i. \label{eq:fim}
\end{equation}

We refer to $A_{i-1}$ and $G_{i}$ as the Kronecker factors.
In practice, $A_{i-1}$ and $G_{i}$ are estimated with a running average of the factors computed over training batches~\cite{martens2015optimizing,pauloski2020convolutional}.

Due to the layer-wise independence of K-FAC, the gradient preconditioning and weight update for a single layer $i$ at iteration $k$ can be written as:
\begin{align}
w_i^{(k+1)}=w_i^{(k)}-\alpha^{(k)}\hat{F}_i^{-1}\nabla L_i(w_i^{(k)}).
\end{align}

We can apply properties~\ref{eq:prop1} and~\ref{eq:prop2} to reduce the gradient preconditioning, $\hat{F}_i^{-1}\nabla L_i(w_i^{(k)})$, to an efficient form where the smaller Kronecker factors, rather than the large FIM, are inverted.
\begin{align}
\hat{F}_i^{-1}\nabla L_i(w_i^{(k)})=G_i^{-1}\nabla L_i(w_i^{(k)})A_{i-1}^{-1}
\end{align}

Tikhonov regularization is used to avoid ill-conditioned matrix inverses with K-FAC by adding a \emph{damping parameter} $\gamma$ to the diagonal of $\hat{F}_i$~\cite{martens2016conv, Osawa_2019_CVPR}.
In most implementations, instead of computing $\hat{F}^{-1}$, we compute
$(\hat{F}_i+\gamma I)^{-1}$ as:
\begin{align}
(\hat{F}_i+\gamma I)^{-1}=(A_{i-1}+\gamma I)^{-1}\otimes (G_i+\gamma I)^{-1}. \label{eq:kfac-damped}
\end{align}

Thus, the final update step for the parameters of layer $i$ at iteration $k$ is:
\begin{align}
w_i^{(k+1)}=&w_i^{(k)}-\alpha^{(k)}(\hat{F}_i+\gamma I)^{-1}\nabla L_i(w_i^{(k)}) \\
=&w_i^{(k)}-\alpha^{(k)}(G_i+\gamma I)^{-1}\nabla L_i(w_i^{(k)})(A_{i-1}+\gamma I)^{-1}.
\end{align}

\subsubsection{Alternative Approximation}

It has been shown that an empirically more stable\footnote{In this context, stable means produces more consistent validation results across batch sizes and hyperparameter settings.} approximation for $(\hat{F}+\gamma I)^{-1}\nabla L_i(w_i^{(k)})$ can be computed using an eigen decomposition of the Kronecker factors, $A$ and $G$, from Equation~\ref{eq:fim}~\cite{martens2016conv,pauloski2020convolutional}. 
Given $Q_A$ and $Q_G$, the eigenvectors of the factors, and $\upsilon_A$ and $\upsilon_G$, the eigenvalues of the factors, the preconditioned gradient can be computed as follows.
\begin{align}
V_1&=Q_G^\top \nabla L_i(w_i^{(k)}) Q_A \label{eq:v1}\\
V_2&=V_1/(\upsilon_G\upsilon_A^\top+\gamma) \label{eq:v2}\\
(\hat{F}_i+\gamma I)^{-1}\nabla L_i(w_i^{(k)})&=Q_GV_2Q_A^\top \label{eq:kfac-eigen}
\end{align}
\revision{The composition of the factors $A$ and $G$ in Equation \ref{eq:fim} guarantees the factors are symmetric and therefore the factor eigen decompostions have real eigenvalues and orthogonal eigenvectors.}
In this work, we use the eigen decomposition method for gradient preconditioning.

\subsubsection{Infrequent K-FAC Updates}
\label{sec:infrequent}

A common strategy in second-order optimization methods is to update the second-order information every few iterations~\cite{martens2015optimizing,Osawa_2019_CVPR,pauloski2020convolutional}.
Intuitively, second-order information does not change as rapidly from one iteration to the next like first-order information.
The K-FAC update interval parameter controls the number of iterations between second-order updates, i.e., iterations between eigen decomposition recomputations.
Larger K-FAC update intervals result in more stale information, so tuning this parameter is key to achieving fast training with K-FAC.
Practically, K-FAC can maintain convergence with K-FAC update intervals of 100--2000 iterations~\cite{pauloski2020convolutional}.

\begin{figure*}[hbt!]
\begin{center}
	\includegraphics[width=0.9\textwidth,trim=0mm 0mm 0mm -6mm, clip]{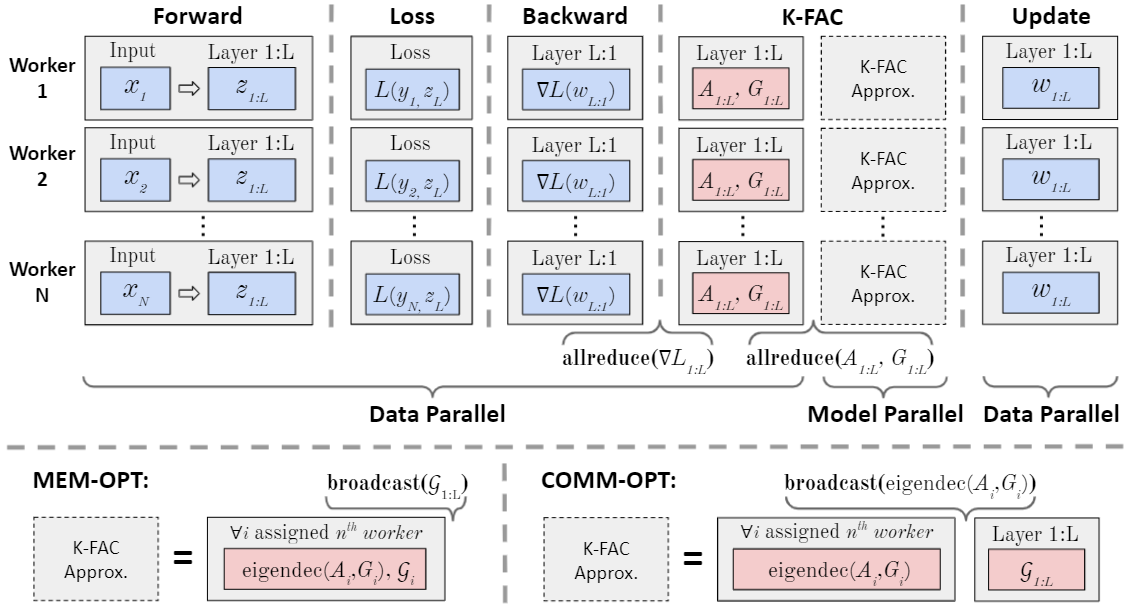}
	\caption{
	    Hybrid-parallel distributed K-FAC implementation overview.
	    Blue boxes are standard computations in data-parallel training.
	    Red boxes are computations required by K-FAC.
	    Workers maintain identical copies of the model.
	    The output of each layer $z$ is computed during the forward pass for the local batch $x$.
	    Then, the loss between the true output $y$ and predicted output $z_L$ is calculated and used to compute gradients in the backward pass.
	    The gradients are then allreduced across workers.
	    During the forward/backward pass, K-FAC caches intermediate data for computing factors.
        In the K-FAC stage, workers compute factors $A$ and $G$ in data-parallel and allreduce the results.
        \revision{
        The eigen decompositions and preconditioned gradients $\mathcal{G}$ are computed in the K-FAC approximation stage.
        The existing K-FAC methods, MEM-OPT and COMM-OPT, implement the model-parallel stage differently as shown at the bottom of the figure.
        After the K-FAC stage, all workers have $\mathcal{G}$ and can update weights locally using $\mathcal{G}$ and a standard optimizer (e.g., SGD or ADAM).
        }
		\label{fig:hybrid-parallel}}
\end{center}
\end{figure*}

\subsection{Distributed Implementation}
Existing distributed K-FAC implementations are hybrid-parallel, with first-order information (e.g., gradients)
computed in data-parallel and second-order information (e.g., K-FAC approximation) in model-parallel.
Figure~\ref{fig:hybrid-parallel} outlines \revision{the model-parallel K-FAC computation performed between standard data-parallel training steps.}

We refer to the two existing distributed K-FAC implementation strategies as MEM-OPT (memory optimized K-FAC) and COMM-OPT (communication optimized K-FAC).
Both work by replicating the DNN across all processes and assigning a random local batch of training data to each process at each iteration.
The data-parallel forward pass, backward pass, and SGD weight update stages outlined in Figure~\ref{fig:hybrid-parallel} are the same in both methods.
The key differences between the two approaches are the model-parallel computation \revision{implementations} in the gradient preconditioning stage.

\subsubsection{MEM-OPT}

MEM-OPT~\cite{Osawa_2019_CVPR} assigns each layer in the DNN to a different process during the preconditioning stage.
Each process computes the eigen decompositions of $A_{i-1}$ and $G_i$ needed for preconditioning the gradients using Equations \ref{eq:v1}--\ref{eq:kfac-eigen}.
Each process then broadcasts the preconditioned gradients 
to all processes such that subsequent SGD weight updates can be done in data-parallel.

MEM-OPT has a low memory footprint because no eigen decompositions are duplicated:
each layer's eigen decompositions are stored on only one process.
MEM-OPT has communication in three places: a) gradient allreduce, b) factor allreduce, and c) preconditioned gradient broadcast, all shown in Figure~\ref{fig:hybrid-parallel}.
In non-K-FAC update steps (e.g., steps where the eigen decompositions are not updated), MEM-OPT avoids the factor allreduce 
because eigen decompositions from a previous step are reused; however, 
the preconditioned gradient broadcast is still required because the gradient being preconditioned changes every iteration.

\subsubsection{COMM-OPT}
COMM-OPT~\cite{pauloski2020convolutional} decouples the eigen decomposition from gradient preconditioning.
Instead of assigning each layer to a process, individual factors are assigned to a process to be eigen decomposed in parallel.
The eigen decompositions are broadcast back to all processes such that every process holds a copy of all eigen decompositions.
Each process computes all preconditioned gradients locally prior to weight updates.

COMM-OPT has a larger memory-footprint because every process must maintain a copy of the eigen decompositions.
COMM-OPT has communication in three places: a) gradient allreduce, b) factor allreduce, and c) eigen decomposition broadcast,
also shown in Figure \ref{fig:hybrid-parallel}.
Decoupling the eigen decompositions from the gradient preconditioning achieves two goals:
1) $A_{i-1}$ and $G_i$ can be computed in different processes, doubling the maximum worker utilization, and
2) in non-K-FAC update steps, the communication in (b) and (c) can be avoided because every worker can locally precondition gradients with the cached eigen decompositions.
Thus, in non-K-FAC update intervals, COMM-OPT has no additional communication overhead compared to SGD.
COMM-OPT has been shown to be 4--16\% faster than MEM-OPT for ResNet-50 training on 16--256 V100 GPUs \cite{pauloski2020convolutional}.

These two implementations convey the fundamental tradeoff between caching the decompositions locally to avoid communication and communicating the preconditioned gradients every iteration to avoid additional memory overheads.
This tradeoff impacts the communication, computation, and memory overhead of K-FAC and motivates our research in this paper.

\section{Design}
\label{sec:design}

We introduce features of \name{}'s design that enable its tunable memory footprint
and explain how \name{} performs load balancing and half precision training to improve scalability.

\subsection{Tunable Memory Footprint}
\label{sec:hybrid-opt}

\begin{figure*}[t]
	\begin{center}
		\centering
		\includegraphics[width=6in]{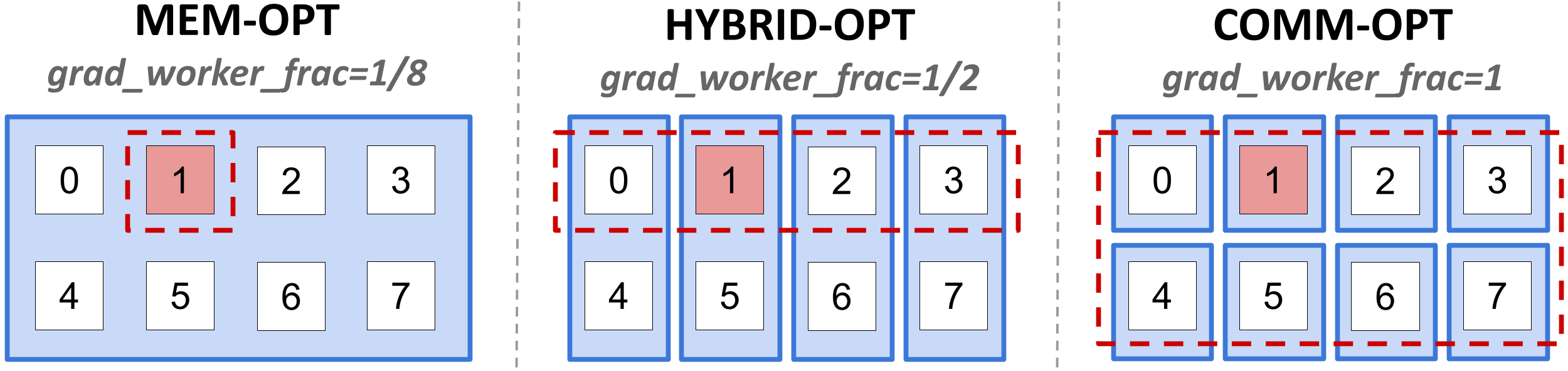}
		\caption{A comparison of three \emph{grad\_worker\_frac} values on eight processes.
			Process 1, shaded in red, computes the eigen decompositions for this layer and communicates the result to all gradient workers, the processes inside the dashed red box.
			Gradient workers compute and broadcast the preconditioned gradient to a subset of the remaining processes, denoted by the light blue box.
			In MEM-OPT, there is one gradient worker that broadcast the preconditioned gradient to all processes.
			In HYBRID-OPT, there are four gradient workers that send the preconditioned gradient to one of the four remaining processes (e.g., process 0 sends the preconditioned gradient to process 4).
			In COMM-OPT, the preconditioned gradient need not be communicated because all processes are gradient workers.
			\label{fig:hybrid-opt}}
	\end{center}
\end{figure*}

KAISA introduces HYBRID-OPT, a configurable distributed K-FAC strategy that can exploit tradeoffs between memory usage and communication overhead.

In this strategy, each layer has a subset of the processes assigned to be gradient preconditioners, referred to as the gradient workers.
\emph{grad\_worker\_frac} defines the size of this subset, i.e., the gradient worker count is $\texttt{max}(1, \textit{grad\_worker\_frac}\times\textit{world\_size})$.
One of the gradient workers is also responsible for computing the eigen decompositions for the layer and broadcasting the results to the remaining gradient workers.
At the end of the eigen decomposition stage, each gradient worker has a copy of the eigen decompositions and can precondition the gradient.

Gradient receivers are the processes not assigned as gradient workers for the layer.
Each gradient worker is responsible for broadcasting the preconditioned gradient to a subset of the gradient receivers.
With multiple gradient workers, the broadcast groups are smaller and simultaneous broadcasts are possible, reducing overall communication time.
For example, given two gradient workers and two gradient receivers, each gradient worker sends the preconditioned gradients to just one receiver; both gradient workers perform this communication at the same time.

After the gradient broadcast, all processes have a copy of the preconditioned gradient with which local weights can be updated.

Observe that \name unifies existing distributed K-FAC strategies because COMM-OPT and MEM-OPT are special cases of HYBRID-OPT. 
COMM-OPT is the case where \emph{grad\_worker\_frac} = 1
(all processes precondition the layer's gradient), and
MEM-OPT is the case where \emph{grad\_worker\_frac} = 1/\emph{world\_size} (a single process preconditions a layer's gradient and broadcasts the result).
Figure \ref{fig:hybrid-opt} compares MEM-OPT, COMM-OPT, and HYBRID-OPT in an eight process environment.

As \emph{grad\_worker\_frac} increases, more processes cache the eigen decompositions and the memory footprint increases.
Visually, the number of processes that cache the eigen decompositions is represented by the processes in the dashed red box in Figure \ref{fig:hybrid-opt}.

Continuing with Figure \ref{fig:hybrid-opt}, HYBRID-OPT has \revision{a lower preconditioned gradient broadcast cost} than MEM-OPT because the broadcast groups are smaller and broadcasts are overlapped.
HYBRID-OPT requires four separate broadcasts to groups of size two in comparison to MEM-OPT which requires one large broadcast to a group of eight.
Since these broadcasts involve non-overlapping processes, we can execute all four broadcast calls simultaneously in the HYBRID-OPT case.
Given the complexity of broadcasting using the minimum spanning tree algorithm is $O(\log p)$, where $p$ is the number of processes in the broadcast group, the complexity is reduced from $O(\log 8)$ in MEM-OPT to $O(\log 2)$ in HYBRID-OPT.
\revision{Note in steps where the eigen decompositions are updated, HYBRID-OPT incurs an additional eigen decomposition broadcast with $O(\log 4)$ complexity; however, as mentioned in \S\ref{sec:infrequent}, eigen decompositions are updated infrequently so the average complexity for HYBRID-OPT is still less than that of MEM-OPT.}

\revision{
The problem addressed by \name's HYBRID-OPT strategy is akin to that of 2.5D matrix multiplication~\cite{edgar2011matrix}.
Both algorithms can utilize extra processor memory to reduce communication costs by controlling the replication factor of data.
In 2.5D matrix multiplication, a parameter $c$ determines the number of data copies, and in \name, the \emph{grad\_worker\_frac} determines the number of workers that store the eigen decompositions.
}

\subsection{Greedy Factor Distribution}
\label{sec:greedy}
The eigen decompositions required for K-FAC are expensive to compute.
Distributed K-FAC optimizes this stage by computing eigen decompositions in a model-parallel fashion.
To most efficiently use available resources, the eigen decomposition computations should be distributed in a manner that minimizes the makespan $T$, the time it takes for all processes to complete their assigned computations.
We use the longest processing time greedy algorithm which produces an assignment with makespan $T\leq \frac{3}{2}T^*$ where $T^*$ is the optimal makespan~\cite{algos2005}.
The longest processing time algorithm sorts jobs by decreasing length and then iteratively assigns each job to the worker with the lowest current work load.

For each factor to be eigen decomposed, the processing time is approximated as $O(N^3)$ where each factor is an $N\times N$ matrix~\cite{demmel2007la}.
Alternatively, memory usage can be optimized for by using $O(N^2)$ as the approximation since $N^2$ is the size of the factor.
\name{} uses the longest processing time strategy at the start of training to assign each factor to a process.

\subsection{Half Precision Storage and Computation}
\label{sec:precision}

Mixed precision training is a common strategy to reduce training times and memory usage on supported hardware~\revision{\cite{micikevicius2018mixed}}.
Popular frameworks for automatic mixed precision (AMP) training, 
such as NVIDIA AMP and PyTorch AMP,
cast forward and backward pass operations to half precision where possible.
Gradient calculations in the backward pass can often produce very small values that would be clipped when cast to half precision, 
so these frameworks scale gradients to a larger value during the backward pass and unscale the gradients appropriately before the optimization step.

\name{} adapts to the precision being used for training.
When training with AMP, factors are stored in half precision, reducing memory costs.
K-FAC computations are performed in half precision where possible.
Eigen decompositions are generally unstable in half precision so factors are cast to single precision before decomposition.
\name{} can store eigen decompositions in half precision to further reduce memory consumption if needed.

Half precision training has become a staple in achieving state-of-the-art results in deep learning, and \name{} can particularly benefit from half precision training due to the K-FAC computation and communication overhead.

\subsection{K-FAC Usage}

\begin{lstlisting}[style=PythonStyle, label={lst:kfac-code}, caption={Example K-FAC usage.}, float, floatplacement=bt]
model = DistributedDataParallel(model)
optimizer = optim.SGD(model.parameters(), ...)
preconditioner = KFAC(model, grad_worker_frac=0.5)

for data, target in train_loader:
    optimizer.zero_grad()
    output = model(data)
    loss = criterion(output, target)
    loss.backward()

    preconditioner.step()
    optimizer.step()
\end{lstlisting}

We design \name{} to implement K-FAC as a preconditioner to standard optimizers with support for Conv2d and Linear layers.
\name{} has an easy-to-use interface and can be incorporated into existing training scripts in two lines: one to initialize and one to call \emph{KFAC.step()} prior to \emph{optimizer.step()} (see Listing \ref{lst:kfac-code}).
\name automatically registers the model and determines the distributed communication backend (e.g., Torch, Horovod, single-process).

A call to \emph{KFAC.step()}: 1) computes the factors using the forward/backward pass data, 2) computes the eigen decompositions in parallel and broadcasts the results, 3) computes the preconditioned gradients and broadcasts the results if necessary, and 4) scales the preconditioned gradients.

\section{Implementation}
\label{sec:impl}

We implement \name{} using  PyTorch~\cite{NEURIPS2019_9015},
with communication, interface, large-batch training, and gradient preconditioning implemented to collectively enable efficient second-order optimization.

\subsection{AMP and Distributed Training}
\label{sec:amp}

We use PyTorch AMP for mixed precision training.
With PyTorch AMP, the \emph{GradScaler} object responsible for scaling and unscaling the gradient in the backward pass can be passed to \name{}.
\name{} uses the \emph{GradScaler} to correctly unscale the $G$ factors, since the scale factor can change from iteration to iteration, causing problems when computing the running average of $G$ over the course of training.
All communication operations are performed in the precision of the data to reduce bandwidth requirements.

\name supports \texttt{torch.distributed} and Horovod~\cite{sergeev2018horovod} for distributed training.
In this work, we use \texttt{torch.distributed} for all experiments because the \texttt{DistributedDataParallel} model wrapper overlaps gradient communication with backpropogation, works seamlessly with PyTorch AMP, and provides a broadcast group abstraction needed for HYBRID-OPT.

\subsection{Factor Accumulation}
\label{sec:factor-acc} 

Processor memory (e.g., GPU VRAM) limits the maximum per-processor batch size during training.
A common strategy to achieve larger effective batch sizes is gradient accumulation, a method where gradient values for multiple forward and backward passes are accumulated between optimization steps.
For example, if processor memory limits the batch size to 8, an effective batch size of 32 can be achieved by accumulating gradients over four forward/backward passes  prior to the optimization step.
This strategy is common in applications such as BERT, where effective batch sizes are often $>2^{15}$, but modern GPUs are limited to local batch sizes $<2^7$.

The forward/backward pass data needed by \name to compute the factors is accumulated over each mini-batch between calls to \emph{KFAC.step()}.
As the number of gradient accumulation steps increases, the memory needed to accumulate the forward/backward pass data grows linearly.
\name can efficiently support gradient accumulation by computing the factors for the current mini-batch during the forward/backward pass instead of during \emph{KFAC.step()}.
The factor communication is still performed during \emph{KFAC.step()}.

\subsection{Triangular Factor Communication}

Previous K-FAC work has exploited the symmetric nature of the Kronecker factors to reduce communication volume by sending only the upper triangle for each factor~\revision{\cite{osawa_2019_icpp, ueno2020kfac}}.
Our implementation supports extracting the upper triangle for the factor allreduce and reconstructing the full factor before the eigen decomposition stage.
For the models studied in this work, this optimization did not yield performance improvements for two reasons.
First, network latency impacted overall communication time more than bandwidth; second, this optimization has an additional overhead for extracting the upper triangle and reconstructing the factor.
For models with larger individual layers---and therefore factors---this optimization could yield greater benefits.

\subsection{Gradient Preconditioning}
\label{sec:preconditioning}

The largest overhead of K-FAC in non-K-FAC update steps is computing the preconditioned gradients.
This process, described by Equations~\ref{eq:v1}--\ref{eq:kfac-eigen}, involves a series of matrix additions, divisions, and multiplications.
Observe that the gradient $\nabla L_i(w_i^{(k)})$ is the only variable that changes between non-K-FAC update iterations.
In particular, the computation involving the outer product of the eigenvalues, $1/(\upsilon_G\upsilon_A^\top+\gamma)$, in Equation~\ref{eq:v2} does not need to be recomputed every iteration---only after the eigen decompositions are updated.
We also observe that when $\emph{grad\_worker\_frac}>1/world\_size$, multiple processes perform this computation redundantly.

To reduce the total number of operations during the preconditioning stage, we move the computation of the outer product into the eigen decomposition stage.
The process assigned to eigen decompose $G$ computes $1/(\upsilon_G\upsilon_A^\top+\gamma)$ and broadcasts the result to all gradient workers instead of broadcasting $\upsilon_A$ and $\upsilon_G$.
This ensures that $1/(\upsilon_G\upsilon_A^\top+\gamma)$ is computed once (on a single worker) and then reused many times by other workers.
In practice, this reduced the time to precondition the gradients for a single layer by up to 53\%.

\section{Experiments}
\label{sec:experiments}

We report on experiments that address four issues:
1) convergence to baseline evaluation metrics; 2) time to convergence with and without \name to validate the design and implementation;
3) exploration of the memory and communication tradeoff using \name{} to develop a quantitative understanding of the 
\textit{grad\_worker\_frac} configuration; and 4) evaluation of \name's scaling performance.

\subsection{Hardware and Software Stack}

We performed experiments on two computers:
\begin{itemize}
    \item The GPU subsystem of the Frontera supercomputer at the Texas Advanced Computing Center, which
    has 112 nodes, each powered by IBM Power9 processors, with four 16 GB NVIDIA V100 GPUs (448 GPUs in total). 
    Nodes are connected by an InfiniBand EDR network. 
    We use PyTorch 1.6, CUDA 10.2, CUDNN 7.6.5, and NCCL 2.5.6, and MVAPICH2-GDR 2.3.4 to launch processes on multiple nodes for distributed training.
    
    \item The GPU subsystem of the Theta supercomputer at Argonne National Laboratory.
    This system has 24 NVIDIA DGXA100 nodes with eight 40GB A100 GPUs each (192 A100 GPUs in total).
    On DGXA100 nodes, we use PyTorch 1.7, CUDA 11.0, CUDNN 8.0.4, and NCCL 2.7.8.
\end{itemize}

We distinguish between these two systems in the following text by specifying either V100 or A100 GPUs.

\subsection{Applications}

We evaluate \name{} for classification, segmentation, and language modeling
applications.

\textbf{Classification:} We use ResNet-50~\cite{he2016deep} with the ImageNet-1k dataset~\cite{krizhevsky2012imagenet},
which has 1000 categories with approximately 1.3M training images and 50K validation images.
We use K-FAC to precondition all convolutional and linear layers in ResNet-50 and SGD for the weight updates.

\textbf{Segmentation:} We explore two segmentation tasks. 
First, we use the NVIDIA reference PyTorch implementation of Mask R-CNN~\cite{he2017mask, nvidiadeeplearning} with the Common Objects in Context (COCO) 2014 dataset~\cite{lin2015microsoft}.
We use K-FAC to precondition the convolutional and linear layers in the region of interest (ROI) heads of Mask R-CNN and SGD for the weight updates.

Second, we use a U-Net~\cite{ronneberger2015u} architecture for segmenting brain tumor sub-regions.
We extend a Kaggle competition implementation~\cite{unetcode} to enable multi-GPU training.
The test case was run on the LGG Segmentation Dataset~\cite{unetdata}, which contains Magnetic Resonance (MR) images
of the brain from 110 patients across five hospitals. 
Images from a random subset of 100 patients are used as the training dataset and the remaining 
10 patients are used for validation.
We apply K-FAC to all convolutional layers in the model.

\textbf{Language Modeling:} We train the BERT-Large Uncased model
using a modified version of the NVIDIA reference PyTorch implementation for BERT~\cite{devlin2018bert, nvidiadeeplearning} with the English Wikipedia~\cite{wiki} and Toronto BookCorpus datasets~\cite{zhu2015aligning}.
Each transformer in BERT is implemented using a series of Linear layers, and we apply K-FAC to all linear layers.
We do not use K-FAC to precondition the embedding layer and prediction head because both of these layers have a Kronecker factor with shape $vocab\_size\times vocab\_size$, and since the vocab size for BERT-Large is 30K, these factors cannot be efficiently eigen decomposed.
Fused LAMB is used as the optimizer~\cite{you2019large}.
The strategy outlined in \S\ref{sec:factor-acc} is used to reduce memory consumption since gradient accumulation is used for training.

\subsection{\revision{Convergence with Fixed Batch Size}}
\label{sec:expe:conv-fixed-batch}
We compare KAISA performance (converged accuracy, epochs to convergence,
and time to convergence) on ResNet-50, Mask R-CNN, U-Net, and BERT-Large against the baseline implementations listed in Table~\ref{tb:baseline}.
For ResNet-50 and Mask R-CNN, we use MLPerf benchmark target results~\cite{mlperf}.
For U-Net, we use the baseline validation Dice similarity coefficient (DSC) from the model's GitHub repository~\cite{unetcode}.
For the BERT-Large baseline, researchers have reported F1 scores of 91.08\%~\cite{nvidiadeeplearning}, 91.0\%~\cite{googlebert}, and 90.4\%~\cite{you2019large} for fine-tuning on the SQuAD v1.1 dataset.
We use the best reported F1 score (i.e., the NVIDIA implementation~\cite{nvidiadeeplearning}) with the LAMB optimizer.
Due to the partial unavailability of the Toronto BookCorpus training dataset, our measurement only converges to 90.8\%.
(The Toronto BookCorpus dataset is no longer available online as a holistic package; 
we could recover only 14,155 of the 16,846 books.)

\begin{table}[]
\begin{center}
    \caption{Baseline performance and hardware summary for ResNet-50, Mask R-CNN, U-Net, and BERT-Large. ``val acc'' is validation accuracy. ``mAP'' is mean average precision. ``DSC'' is Dice similarity coefficient.
    }
    \begin{small}
    \begin{tabular}{ | c | c | c | c | c |}
    \hline
    App            & Ref & Baseline   & GPU & \# GPUs \\ \hline \hline
    \multirow{2}{*}{ResNet-50} & \multirow{2}{*}{\cite{mlperf}} & \multirow{2}{*}{75.9\% val acc} & V100 & 64\\ \cline{4-5}
    & & & A100 & 8 \\ \hline
    \multirow{2}{*}{Mask R-CNN}  & \multirow{2}{*}{\cite{mlperf}} &  0.377 bbox mAP, & \multirow{2}{*}{V100} & 32 \\ \cline{5-5}
    & & 0.342 segm mAP &  &  64 \\ \hline
    U-Net & \multirow{1}{*}{\cite{unetcode}} & 91.0\% val DSC & A100 & 4 \\ \hline
    \multirow{2}{*}{BERT-Large} & \multirow{2}{*}{\cite{nvidiadeeplearning}} & 90.8\% SQuAD   & \multirow{2}{*}{A100} & \multirow{2}{*}{8}\\ 
     &  & v1.1 F1 score  &  & \\ \hline

    \end{tabular}
    \end{small}
    \label{tb:baseline}\up\up
\end{center}   
\end{table}

\revision{In all comparisons, we use the same global batch size for KAISA and the original optimizers to isolate the improvement from second-order information.}
Table~\ref{tb:hyper} summarizes the hyperparameters for each application.
\revision{
For ResNet and K-FAC specific hyperparameters, we use values from~\cite{pauloski2020convolutional} to provide direct performance comparisons to previous K-FAC works.
For Mask R-CNN and BERT-Large, we use the NVIDIA reference hyperparameters~\cite{nvidiadeeplearning}. 
Further performance improvements could be gained through more extensive hyperparameter tuning; however, this was not necessary to achieve results  better than the original optimizers.
}

\begin{table}[]
\begin{center}
    \caption{Summary of hyperparameters used for each application. BS = global batch size, LR = learning rate, WU = warm up iterations, K\_freq = number of iterations between eigen decomposition re-computations, F\_freq = iterations between factor updates. \emph{grad\_worker\_frac} = 1 and damping = 0.003 for all cases.}
    
    \begin{small}
    \begin{tabular}{ | c | c | c | c | c | c |}
    \hline
    App   & BS & LR   & WU & K\_freq & F\_freq \\ \hline \hline
    ResNet-50 & 2,048 & 0.8 & 3,130 & 500 & 50 \\ \hline
    Mask R-CNN & 64  & 8e-2 & 800 & 500 & 50  \\ \hline
    U-Net & 64 & 4e-4& 500 & 200 & 20 \\ \hline
    BERT-Large & 65,536  & 5e-5 & 103 & 100 & 10  \\ \hline
    \end{tabular}
    \end{small}
    \label{tb:hyper}\up\up
\end{center}   
\end{table}

\begin{figure*}[]
  \subfigcapmargin=0.1in
  \subfigure[ResNet-50. The time-to-convergence is 268.1 mins for K-FAC and 354.0 mins for momentum SGD.]{
    \captionsetup{width=.1\linewidth}
    \includegraphics[width=2.25in,trim=2mm 2mm 2mm 2mm,clip]{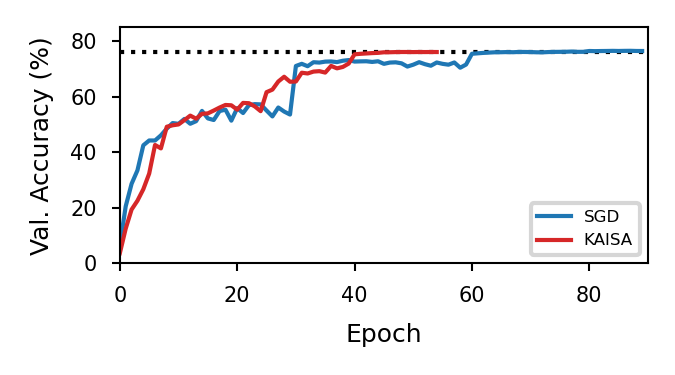}%
    \label{fig:conv:resnet}
  }
  \subfigure[Mask R-CNN. The time-to-convergence is 115.8 mins for K-FAC and 136.1 for SGD.]{
    \includegraphics[width=2.25in,trim=2mm 2mm 2mm 2mm,clip]{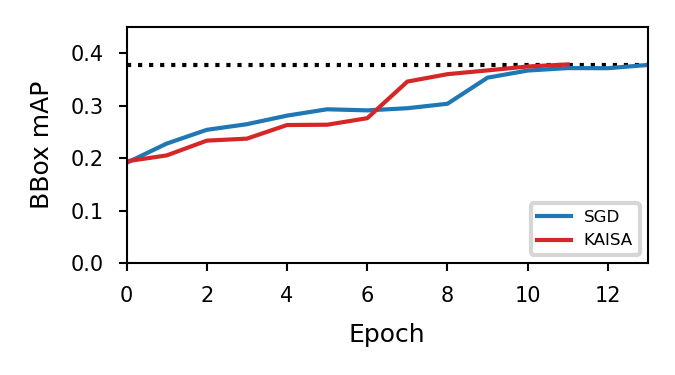}%
    \label{fig:conv:maskrcnn}
  }
  \subfigure[U-Net. The time-to-convergence is 10.9 mins for K-FAC and 14.6 for ADAM.]{
    \includegraphics[width=2.25in,trim=2mm 2mm 2mm 2mm,clip]{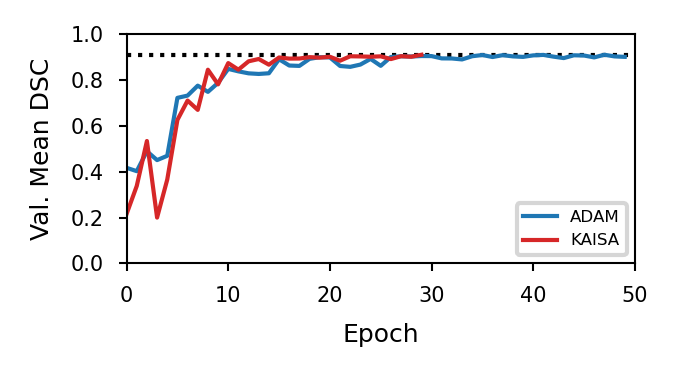}%
    \label{fig:conv:unet}
  }
  \caption{Validation Metric Curve Comparison between \name{} and SGD/ADAM on ResNet-50, Mask R-CNN, and U-Net. The dotted lines represent the target metric.}
  \label{fig:convergence}
\end{figure*}

\textbf{ResNet-50:}
We train ResNet-50 for 55 and 90 epochs for \name{} and momentum SGD, respectively, using FP16 precision on eight NVIDIA A100 GPUs.
Figure~\ref{fig:conv:resnet} shows the validation accuracy curve using the two optimization methods.
\name{} converges to the baseline validation accuracy at epoch 46 and momentum SGD at epoch 65.
The time-to-convergence is 268.1 minutes for \name{}: 24.3\% less than the 354.0 minutes for momentum SGD.

\subfigcapmargin = 0pt

\textbf{Mask R-CNN:}
Our baseline measurement of Mask R-CNN on 32 NVIDIA V100 GPUs takes 25,640 iterations to converge to 0.377 bbox mAP and 0.342 segm mAP in FP32.
With K-FAC enabled for the ROI heads, training converges to 0.379 bbox mAP and 0.350 segm mAP in 21,000 iterations.
The bbox mAP comparision for SGD and \name{} is shown in Figure~\ref{fig:conv:maskrcnn}.
\name{} reduces the training time from 136.1 minutes to 115.8 minutes, a 14.9\% improvement.
\revision{With a batch size of 128 on 64 V100 GPUs,} \name{} converges in 12,000 iterations compared to 15,000 with SGD and reduces training time by 18.1\%.

\textbf{U-Net:}
The reference U-Net implementation~\cite{unetcode} with ADAM converges to 91.0\% validation DSC within 50 epochs with four NVIDIA A100 GPUs using FP32 training.
Using \name{}, the training converges above 91.0\% in 30 epochs as seen in Figure~\ref{fig:conv:unet}.
\name{} reduces the training time by 25.4\% (10.9 minutes vs. 14.6 minutes).

\textbf{BERT:}
\revision{BERT pretraining has two phases}.
The first trains with maximum sequence length of 128 for 7038 iterations and the second trains with maximum sequence length of 512 for 1,563 iterations.
We fine-tune the pretrained BERT model for three epochs using the SQuAD dataset.
All phases and fine-tuning are done in FP16.
Tuning the hyperparameters of BERT is expensive as each run can take over 130~hours with 8 A100 GPUs,
so we showcase the effectiveness of \name{} with the second phase of BERT pretraining.
\revision{For phase two pretraining, we start with the same model pretrained with LAMB during phase one.}
We train with \name{} for \{800, 1,000, 1,200\} iterations then fine-tune SQuAD.
Table~\ref{tb:BERT} summarizes the validation SQuAD F1 scores after fine-tuning and the pretraining performance improvements.

\begin{table}[]
\begin{center}
    \caption{BERT performance comparison: \name{} vs.\ LAMB}
    \begin{small}
    \begin{tabular}{ | c | c | c | c | c |}
    \cline{3-5}
    \multicolumn{2}{}{} & \multicolumn{3}{|c|}{\name{}, with iterations:}\\ \hline
    Metric & LAMB  &  1200 &  1000 & 800\\ \hline \hline
    F1 & 90.8 & 91.0 & 91.0 & 90.8\\ \hline
    Stdev  & 0.13 & 0.15 & 0.17 & 0.24\\ \hline
    Iterations & 1,536 & 1,200 & 1,000 & 800 \\ \hline
    Time (hrs) & 47.7 & 41.5  & 34.4 & 30.4\\ \hline
    \end{tabular}
    \end{small}
    \label{tb:BERT}
\end{center}   
\end{table}

\name{} converges to the 90.8 F1 baseline in 800 iterations, 47.9\% less iterations than required for LAMB, and \name{} takes 36.3\% less time than LAMB to converge.

\subsection{\revision{Convergence with Fixed Memory Budget}}
\revision{
To understand how a \textit{grad\_worker\_frac} can enable second-order optimization in memory-constrained environments, we compare \name's performance against baselines with fixed memory budgets.
In particular, we train ResNet-50 on 64 V100 GPUs and BERT-Large phase two on eight A100 GPUs. 
For each experiment, we use the maximum possible local batch size and measure the convergence, epochs to convergence, and time to convergence.
Table~\ref{tb:hyper-mem-budget} summarizes the hyperparameters and time to convergence.}

\begin{table}[]
\begin{center}
    \caption{\revision{Time to convergence and hyperparameters for each optimizer. BS = global batch size, LR = learning rate, g\_frac = gradient worker fraction, K\_f = iterations between eigen decomposition re-computations, F\_f = iterations between factor updates, T\_conv is time to convergence in minutes.}}
    
    \begin{small}
    \begin{tabular}{ | c | c | c | c | c | c | c | c |}
    \hline
    App & Opt   & BS & LR   & g\_frac & K\_f & F\_f & T\_conv \\ \hline \hline
    \multirow{3}{*}{ResNet-50} & SGD & 8K & 3.2 & --- & --- & --- & N/A  \\ \cline{2-8}
    & KAISA & 5K  & 2.0 & 1/64 & 200 & 20 & 96   \\ \cline{2-8}
    & KAISA & 5K  & 2.0 & 1/2 & 200 & 20 & 83   \\ \hline
    \multirow{3}{*}{BERT-Large} & LAMB & 24K & 3e-3 & --- & --- & --- & 2,917.6  \\ \cline{2-8}
    & KAISA & 32K  & 4e-3 & 1/2 & 100 & 10 & 1,702.5  \\ \cline{2-8}
    & KAISA & 32K  & 4e-3 & 1 & 100 & 10 & 1,703.5  \\ \hline
    \end{tabular}
    \end{small}
    \label{tb:hyper-mem-budget}
\end{center}   
\end{table}

\revision{\textbf{ResNet-50:}
With momentum SGD, the maximum local batch size is 128 per GPU and the global batch size is 8,192.
We train ResNet-50 for 90 epochs using momentum SGD which achieves 75.0\% validation accuracy---0.9\% lower than the MLPerf baseline. 
Next, we use KAISA with \textit{grad\_worker\_frac} = 1 and a local batch size of 80; however, the training runs out of memory.
Lowering \textit{grad\_worker\_frac} to 1/2, training takes 83 minutes to converge to 75.9\% in the 48th epoch. 
The complete 55 epoch training takes 95 minutes and the validation accuracy reaches 76.0\%. 
So, even if momentum SGD converges to 75.9\% by the 90th epoch, KAISA still reduces the time to convergence by 32.5\%. 
Finally, we run the same training process with MEM\_OPT by setting \textit{grad\_worker\_frac} to 1/64.
It converges at the 47th epoch and the time to convergence is 96 minutes.
The complete 55 epoch training takes 111 minutes. 
This experiment highlights the benefit of KAISA in cases where compute resource can not be efficiently utilized with the original optimizers.
With a \textit{grad\_worker\_frac} value of 1/2, KAISA offers benefits over COMM\_OPT and MEM\_OPT by enabling second-order optimization under a tight memory budget that is 13.9\% faster than COMM\_OPT and not feasible with MEM\_OPT.}

\revision{\textbf{BERT:}
With LAMB, the maximum possible local batch size per GPU is 12 for the second phase of this BERT implementation~\cite{nvidiadeeplearning}, and the global batch size is 24,576. 
For KAISA, we use a local batch size of 8 and global batch size of 32,768. 
This experiment uses the same hyperparameters as in \S\ref{sec:expe:conv-fixed-batch}, so all cases with LAMB and BERT should converge to the baseline, thus we only project the training time with the first 100 steps.
As the global batch size with LAMB is 24,576, 2,084 training steps are required to finish three epochs, and the training time is 2,917.6 minutes.
KAISA, with \textit{grad\_worker\_frac} = 1/2, takes 3,268.8 minutes to finish the three epochs. 
However, KAISA converges to baseline after 800 steps, as shown in Table~\ref{tb:BERT},
so the time to converge is 1,702.5 minutes---41.6\% faster than LAMB.
Setting \textit{grad\_worker\_frac} = 1 takes 1,703.5 minutes to converge for KAISA.
The performance is comparable to the case with \textit{grad\_worker\_frac} = 1/2. 
We conduct a detailed study on this convergences phenomena for different \textit{grad\_worker\_frac} values in \S\ref{sec:mem-vs-comm}.
}

\subsection{Memory vs.\ Communication}
\label{sec:mem-vs-comm}

\begin{figure*}[bt]
    \centering
	\subfigure[\revision{ResNet-18 (FP32)}]{
		\includegraphics[width=2.25in,trim=2mm 2mm 2mm 1mm,clip]{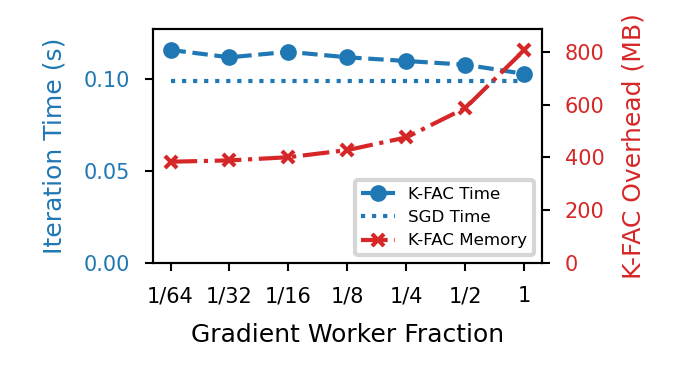}%
	}
	\subfigure[\revision{ResNet-50 (FP32)}]{
		\includegraphics[width=2.25in,trim=2mm 2mm 2mm 1mm,clip]{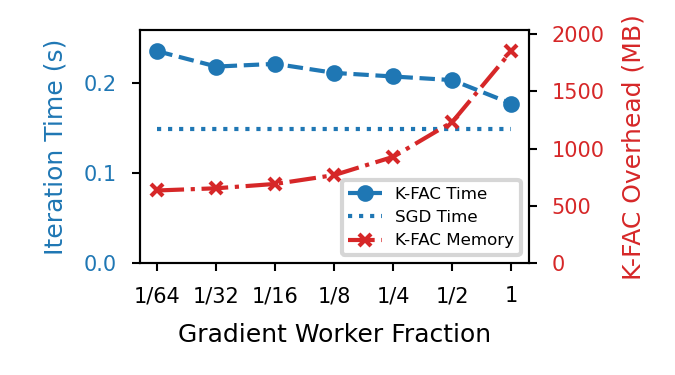}%
	}
	\subfigure[\revision{ResNet-101 (FP32)}]{
		\includegraphics[width=2.25in,trim=2mm 2mm 2mm 1mm,clip]{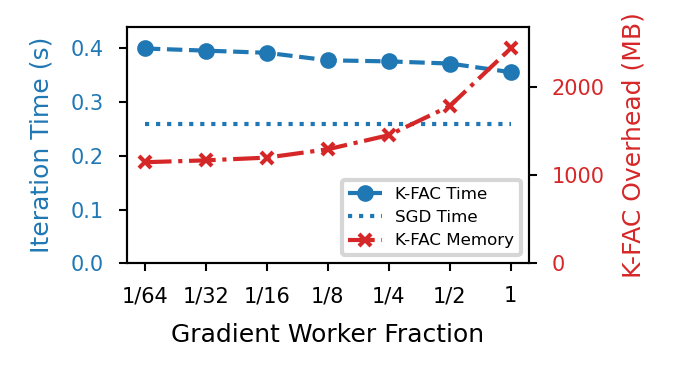}%
	}
	\\
	\subfigure[\revision{ResNet-152 (FP32)}]{
		\includegraphics[width=2.25in,trim=2mm 2mm 2mm 1mm,clip]{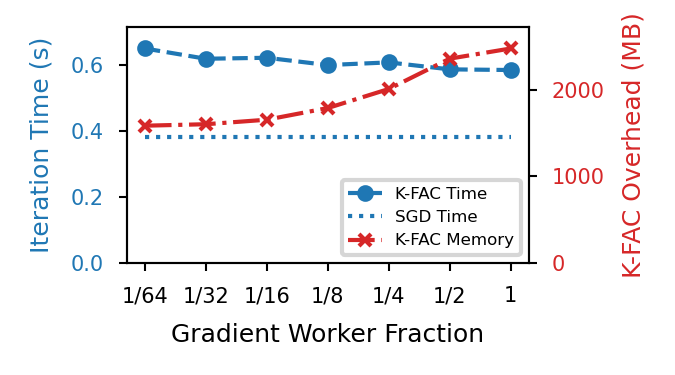}%
	}
	\subfigure[\revision{Mask R-CNN (FP32)}]{
		\includegraphics[width=2.25in,trim=2mm 2mm 2mm 1mm,clip]{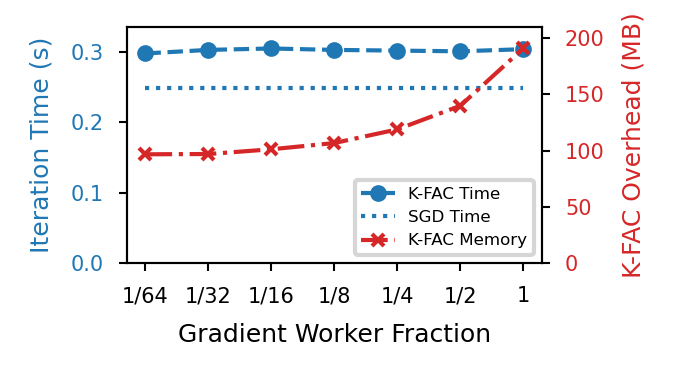}%
	}
	\subfigure[\revision{BERT-Large Phase 2 (FP16)}]{
		\includegraphics[width=2.25in,trim=2mm 2mm 2mm 2mm,clip]{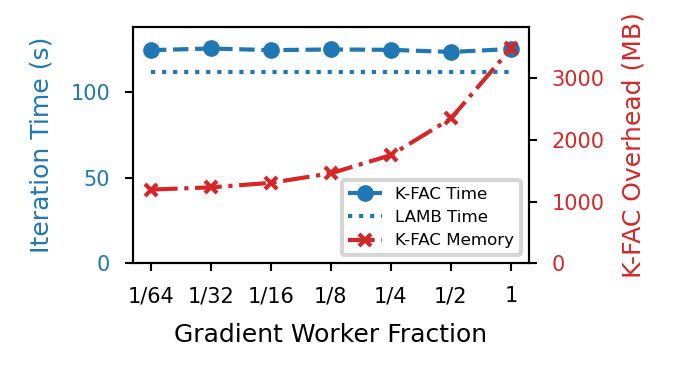}%
	}
	\caption{Average iteration time and K-FAC memory overhead across \emph{grad\_worker\_frac} values on 64 V100 GPUs.
	\revision{Dotted lines are the baseline iteration times without K-FAC.}}
	\label{fig:grad-frac}
\end{figure*}
 
To understand the impact of \emph{grad\_worker\_frac} on training times, we train \revision{ResNet-\{18, 50, 101, 152\}}, Mask R-CNN, and BERT-Large on 64 V100 GPUs for \emph{grad\_worker\_frac} $\in$ \{1/64, 1/32, 1/16, 1/8, 1/4, 1/2, 1\}.
For each experiment, we record the average iteration time, i.e., time between weight updates, and the GPU memory usage.
We refer to the \emph{K-FAC overhead} as the memory required to store the factors and eigen decompositions, and the K-FAC overhead is computed as the difference between the K-FAC and SGD memory usage.
\revision{For all ResNet models, we use the same hyperparameters used for ResNet-50 (Section~\ref{sec:expe:conv-fixed-batch}) except for ResNet-152 where the local batch size was lowered to 24.}
The results are presented in Figure~\ref{fig:grad-frac}, and a summary of the memory usage is provided in Table~\ref{tab:mem-usage}.

The K-FAC memory overhead increases linearly as a function of \emph{grad\_worker\_frac} for all models.
\name requires 1.5--45.8\% more memory than SGD depending on the application and value of \emph{grad\_worker\_frac} (Table~\ref{tab:mem-usage}).
The maximum K-FAC overhead (i.e., when \emph{grad\_worker\_frac} = 1) is 1.5--2.9$\times$ that of the minimum K-FAC overhead (i.e., when \emph{grad\_worker\_frac} = $1/64$).

\begin{table}[]
\begin{center}
	\caption{Summary of per-GPU memory usage for training, in MB. 
	Abs. is the absolute memory required for training.
	$\Delta$ is the \%-increase in memory required over SGD.
	The K-FAC overhead is the K-FAC abs. memory minus the SGD abs. memory.
	}
	\begin{small}
		\begin{tabular}{|c|c||c|c|c|c|c|}
			\cline{3-7}
			\multicolumn{2}{c|}{} & SGD  & \multicolumn{2}{c|}{K-FAC Min} & \multicolumn{2}{c|}{K-FAC Max} \\ \hline
			Model & Precision    & Abs. & Abs. & $\Delta$ & Abs.       & $\Delta$ \\ \hline \hline
			\revision{ResNet-18}            & FP32 & 2454 & 2838 & 16.7\% & 3260  & 32.8\% \\ \hline
			ResNet-50            & FP32 & 4762 & 5396 & 13.3\% & 6608  & 38.8\% \\ \hline
			\revision{ResNet-101}           & FP32 & 6313 & 7463 & 18.2\% & 8755  & 38.7\% \\ \hline
			\revision{ResNet-152}           & FP32 & 6620 & 8204 & 23.9\% & 9092  & 37.3\% \\ \hline
			Mask R-CNN           & FP32 & 6553 & 6650 & 1.5\%  & 6743  & 2.9\%  \\ \hline
			BERT-Large           & FP16 & 8254 & 9555 & 15.8\% & 12038 & 45.8\% \\ \hline
		\end{tabular}
	\end{small}
	\label{tab:mem-usage}
\end{center}
\end{table}

\revision{With respect to iteration times, the ResNet models scale well with the number of gradient workers with ResNet-50 scaling the best.
For ResNet-50, the speedup from a gradient worker count of 1 to 64 is 24.4\% for FP32 with a 22\% increase in total memory usage.}
The average iteration times for Mask R-CNN and BERT-Large remain constant as the number of gradient workers is increased.

For comparison, the same ResNet-50 experiment with 64 V100s in \cite{pauloski2020convolutional} only shows a 7.6\% speedup when increasing the gradient worker count from 1 to 64.
This improvement in KAISA over previous work is due to the unique contributions presented in \S\ref{sec:greedy}, \S\ref{sec:precision}, \S\ref{sec:amp}, and \S\ref{sec:preconditioning}.
These promising results for ResNet-50, a de facto standard benchmark for deep learning systems, are important as the performance characteristics of ResNet-50 represent a large set of commonly used models (e.g. VGG16, U-Nets, etc.).

We can understand why ResNet \revision{model} performance varies across \emph{grad\_worker\_frac} values while the Mask R-CNN and BERT-Large performance  remains constant by considering the bandwidth requirements of these applications.
The bandwidth required by \name{} is a function of the size of the factors, eigen decompositions, and frequency of K-FAC updates.

\revision{With 64 V100s, ResNet-50 calls \emph{KFAC.step()} frequently (4--6 calls/second) and incurs a K-FAC memory overhead between 634~MB and 1.8~GB.}
In comparison, Mask R-CNN calls \emph{KFAC.step()} with a lower frequency (3 calls/second) and has a much smaller K-FAC overhead (100--200~MB).
Thus, the changes in how \name{} communicates data with respect to \emph{grad\_worker\_frac} are less apparent in Mask R-CNN.
BERT-Large has the lowest bandwidth requirements of all applications even though it has the largest K-FAC overhead.
BERT-Large uses gradient accumulation to achieve very large batch sizes \revision{(32K for phase 2)} and as a result only calls \emph{KFAC.step()} every \revision{$\sim$120} seconds.

While the iteration times for low-communication models such as BERT and Mask R-CNN are invariant to the \emph{grad\_worker\_frac} value in \name,
\name still produces faster-than-SGD training times with small increases in memory-overhead. 
This is due to \name's unique features outlined in \S\ref{sec:design} and \S\ref{sec:impl}.
Further, practitioners training these models at larger scales, e.g., 100s or 1000s of GPUs, where communication becomes a greater bottleneck will benefit more from the flexibility \name provides to adapt training to environments with increasing communication costs.

Tuning the \emph{grad\_worker\_frac} hyperparameter, to determine an optimal balance between iteration time and memory usage, is simple as it only requires profiling the average iteration time for each \emph{grad\_worker\_frac} value over a few iterations.

\begin{figure}[]
	\includegraphics[width=\columnwidth,trim=2mm 2mm 2mm 2mm,clip]{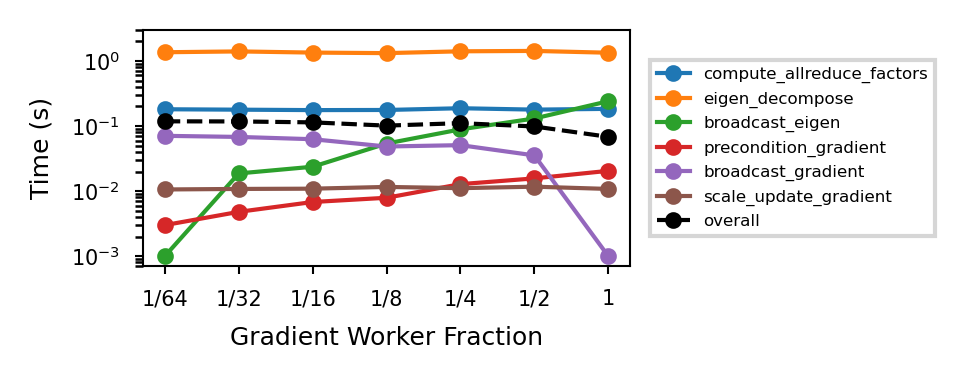}%
	\caption{Average function execution time during calls to \emph{KFAC.step()} for ResNet-50 on 64 GPUs.}
	\label{fig:kfac-trace}
\end{figure}

We find that with respect to training times, \emph{grad\_worker\_frac} has the most impact in applications that spend a larger proportion of time doing communication.
To further understand how \emph{grad\_worker\_frac} impacts training times, we analyze the execution time for each section within \emph{KFAC.step()} with ResNet-50 on 64 V100s.
\revision{
Figure \ref{fig:kfac-trace} provides the time spent in each section for all layers in the model during a call to \emph{KFAC.step()}.
Times are averaged over 10,000 iterations and across all workers.
}
Eigen decompositions are updated every 500 iterations.
As shown in Figure \ref{fig:kfac-trace}, factor computation and communication, eigen decomposition, and scaling and updating the gradients, are invariant to the \emph{grad\_worker\_frac}.

The time required to broadcast the eigen decompositions increases substantially as the number of gradient workers is increased.
Referring back to Figure \ref{fig:hybrid-opt}, the more gradient workers there are, the more processes that need to receive the eigen decompositions.
However, the eigen decompositions, in this case, are only recomputed every 500 iterations so the eigen decomposition broadcast has a negligible effect on the average iteration time.

On the other hand, the gradient preconditioning and broadcast occur every iteration regardless of if the factors or eigen decompositons are updated and have the greatest influence on average iteration time.
We see that the time to precondition the gradients increases with the gradient worker count because each process is assigned as a gradient worker for more layers.
This discovery highlights the importance of the gradient preconditioning optimizations made in \S\ref{sec:preconditioning}.
The time to broadcast the preconditioned gradients decreases to 0 as the gradient worker count approaches \emph{world\_size}, and notably, the time decreases at a \emph{faster rate} than the increase in time required in the preconditioning stage.
This trend is a result of each gradient worker needing to send the results to fewer other processes as the gradient worker count increases.

\subsection{Scaling}

\begin{figure}[]
    \centering
	\subfigure[ResNet-50]{
		\includegraphics[width=\columnwidth,trim=2mm 2mm 0mm 1mm,clip]{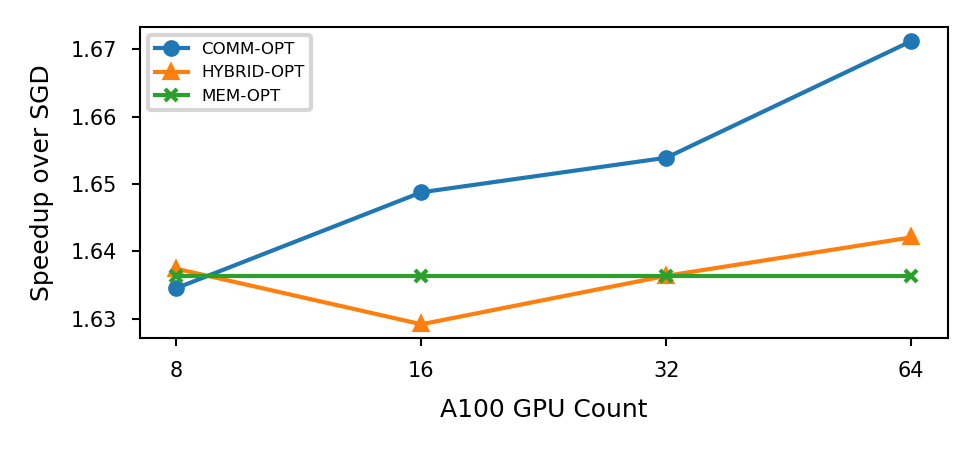}%
		\label{fig:scaling-1}
	}
	\subfigure[BERT-Large Phase 2]{
		\includegraphics[width=\columnwidth,trim=2mm 2mm 0mm 1mm,clip]{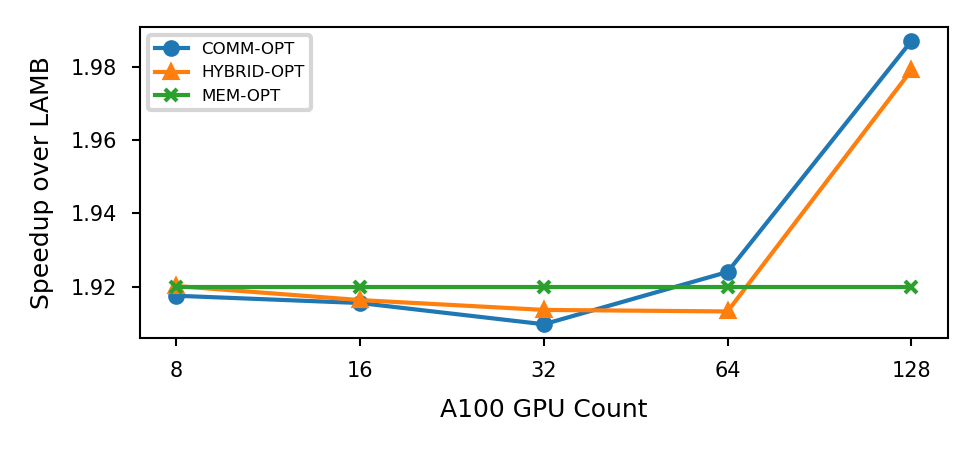}%
		\label{fig:scaling-2}
	}
	\caption{Speedup for the ResNet-50 and BERT-Large applications on A100 nodes.}
	\label{fig:scaling}
\end{figure}

\revision{
To examine the scaling characteristic of \name, we measure the average time per epoch for ResNet-50 and average time per iteration for BERT-Large phase 2.
We choose ResNet-50 and BERT-Large because they represent a high-communication and low-communication model, respectively (see \S\ref{sec:mem-vs-comm}).
We study three \name variants: COMM-OPT, MEM-OPT, and HYBRID-OPT with a \emph{grad\_worker\_frac} of 1/2, and we report the projected end-to-end training time speedup for the \name variants over the base optimizers (SGD and LAMB) in Figure~\ref{fig:scaling}.
For ResNet-50, we project the training time to 90 epochs for SGD and 55 for \name.
For BERT-Large, we project the phase 2 training time to 1,563 steps for LAMB and 800 for \name based on the study in \S\ref{sec:expe:conv-fixed-batch}.
The hyperparameters in Table~\ref{tb:hyper} are used, and for ResNet-50, the K-FAC update frequency is scaled inversely with the global batch size to keep the number of K-FAC updates per training samples constant.
We store and communicate the factors and eigen decompositions in FP16 for BERT-Large.
}

\revision{
In Figure \ref{fig:scaling-1} and \ref{fig:scaling-2}, MEM-OPT maintains a constant speedup over SGD and LAMB, respectively, across all scales.
In contrast, the speedup for COMM-OPT improves in both cases as the scale increases indicating the tradeoff of more memory for reduced communication does have scaling benefits.
HYBRID-OPT sees performance improvements on par with COMM-OPT with BERT-Large while using less GPU memory. Overall, HYBRID-OPT has the best balance of scaling and memory usage for large-scale BERT pretraining.
}

\revision{
As a whole, \name achieves better speedups with BERT-Large which is likely due to ResNet-50 being more communication bound than BERT-Large as noted in \S\ref{sec:mem-vs-comm}.
Further scaling experiments on a machine with more GPUs is needed to understand the characteristics and limits of \name's speedup over SGD.
}

\section{Related Work}
\label{sec:related}

The emergence of DNNs has motivated revisiting many aspects of system design.
TensorFlow~\cite{Abadi2016}, PyTorch~\cite{NEURIPS2019_9015}, and MXNet~\cite{Chen2015} are examples of deep learning frameworks that support end-to-end training.
TensorFlow Serving~\cite{olston2017tensorflow}, DLHub~\cite{chard19dlhub}, and Clipper~\cite{crankshaw2017clipper} focus on low-latency inference serving.
Ray~\cite{moritz2018ray} provides a platform to integrate simulation, training, and serving for reinforcement learning applications.
Researchers have also designed new scheduling approaches in GPU clusters with informed hardware heterogeneity~\cite{narayanan2020heterogeneity}, sharing capability~\cite{xiao2020antman}, and early user feedback~\cite{xiao2018gandiva} to optimize cost and hardware utilization.
Our work here focuses on a lower-level question: given model architecture, communication bandwidth, processor count, and memory size, 
\revision{can the hybrid-parallel aspect of distributed K-FAC be adaptable and reduce training time?}
In the rest of this section, we briefly review other works in optimization frameworks.

{\bf Optimizer Frameworks:}
Distributed optimization frameworks take many forms.
In a synchronous optimizer such as Horovod~\cite{sergeev2018horovod},
all variables are updated in every iteration.
Asynchronous optimizers relax variable update consistency,
for example by passing values via parameter servers~\cite{li2014scaling},
to achieve higher performance than synchronous methods~\cite{recht2011hogwild, lin2017deep}.
The pipeline parallel paradigm, such as GPipe~\cite{huang2019gpipe} and PipeDream~\cite{narayanan2019pipedream} which hold multiple versions of a model partition in a processor and exploit asynchronous optimizers, has shown comparable training convergence.
BytePS~\cite{jiang2020unified} proposes a unified interface for synchronous and asynchronous SGD.
Generally, asynchronous SGD has a non-linear slowdown compared to synchronous SGD~\cite{alistarh2018convergence}.
\name{} implements K-FAC in a synchronous manner as a preconditioner such that it can work with any SGD variant.

{\bf Memory Shortage and Remedies:}
Training DNNs is memory intensive. A common technique
for training large models is to swap between processor memory and host memory, such as in  SwapAdvisor~\cite{huang2020swapadvisor} and SuperNeurons~\cite{wang2018superneurons}.
An alternative is to discard some activation tensors and rematerialize them when needed for back-propagation.
Checkmate~\cite{jain2020checkmate} formulates this tradeoff as an optimization problem and provides an optimal rematerialization schedule.
In \name{}, we apply techniques such as precision relaxing to reduce the memory consumption of K-FAC
and controlling the distribution of eigen decomposition results across processors to maintain a minimal cost in training time.

{\bf K-FAC Convergence:}
Prior work on distributed K-FAC has reported training convergence primarily on ResNet-like convolutional neural networks.
One study~\revision{\cite{Ba2017Distributed}} used asynchronous distributed K-FAC to train ResNet-50 with ImageNet 2$\times$ faster than standard SGD, but only achieved 70\% validation accuracy, a 5.9\% loss compared to the 75.9\% MLPerf~\cite{mlperf} baseline.

Another distributed K-FAC implementation~\cite{Osawa_2019_CVPR} trained ResNet-50 with ImageNet to 74.9\% validation accuracy in just 978 iterations; however, comparisons to SGD are not provided.
\revision{Later work iterated on this implementation to achieve 75.0\% validation accuracy on ImageNet with ResNet-50 on 2048 V100 GPUs in 2 minutes by carefully optimizing the baseline SGD training and introducing a 21-bit floating point (FP21) specification for the factors along with other optimizations from previous works such as triangular matrix communication and infrequent K-FAC updates.
FP21 was introduced due to worse convergence when using FP16 for factor communication; however, with \name{}, we found similar validation accuracy with ResNet-50 for FP32 and FP16 factor communication.
Differences in the numerical stability of the eigen decomposition in \name{} rather than matrix inversion could be a possible reason for the discrepancies.
}

A fourth study~\revision{\cite{pauloski2020convolutional}} trained ResNet-50 to the 75.9\% MLPerf baseline in 18-25\% less time than with SGD by replacing the factor inverse with eigen decomposition and optimizing for reduced communication in non-K-FAC update steps, at the cost of a high memory footprint.
\name{} generalizes previous distributed K-FAC strategies via a tunable memory footprint to balance the memory and communication costs.
This design enables efficient, distributed K-FAC research across a wider range of hardware.
Further, we showcase \name{}'s effectiveness on a variety of domains and models.

{\bf Alternative K-FAC Methods:}
Eigenvalue-corrected Kronecker-factorization (EK-FAC)~\cite{ekfac2018}, a more accurate approximation of the FIM, can perform cheap, partial updates.
Noisy K-FAC~\cite{noisykfac2018l} and noisy EK-FAC~\cite{noisyekfac2018} are functionally similar to standard K-FAC but introduce adaptive weight noise.
K-BFGS and K-BFGS(L)~\cite{quasi-newton2020} apply BFGS~\cite{bfgs70} and L-BFGS~\cite{lbfgs89} in an analogous method to K-FAC (e.g., block-diagonal, Kronecker-factored).
These works have shown better-than-K-FAC performance in many small scale (e.g., single GPU) and small dataset/model (e.g., MNIST or CIFAR-10 with VGG16) cases.
Kronecker-factor based FIM approximation variants are a growing area of research; however, large-scale studies have largely been limited to standard K-FAC. 
\name{} introduces a valuable, unified design paradigm that can be applied to these K-FAC variants to efficiently deploy and evaluate their effectiveness on large models at scale.

\section{Conclusion}
\label{sec:conc}

We have presented \name{}, a {\bf K}-FAC-enabled, {\bf A}daptable, {\bf I}mproved, and {\bf S}c{\bf A}lable second-order optimizer framework.
To enable scalable DNN training, \name{} adapts memory and communication usage, and appropriately distributes the complex K-FAC computations to best suit the model and hardware characteristics.
We design \name to be adaptable to hardware with limited memory (e.g., gaming GPUs) or environments with high communication costs (e.g., Ethernet or massively parallel).
We study the fundamental tradeoff between data access on local and remote memory, and evaluate KAISA's correctness and impact on training time by using four real-world applications.
\revision{With the same global batch size, our experiments show a 18.1--36.3\% training time reduction for ResNet-50, Mask R-CNN, U-Net, and BERT-Large, while preserving convergence to the baseline.
Under the same memory budget, ResNet-50 and BERT phase 2 converge to the baseline in 32.5\% and 41.6\% less time compared to momentum SGD and Fused LAMB. 
In high-communication applications, such as ResNet models, we show that extra processor memory can be used to improve iteration times by reducing communication.
In low-communication applications, such as Mask R-CNN and BERT-Large, we show the optimal \name usage and efficient scaling on par with SGD up to 128 GPUs.
}

\section{Acknowledgements}

This work was supported by NSF OAC-1931354 and OAC-1818253. 
Shivaram Venkataraman is supported by the Office of the Vice Chancellor for Research and Graduate Education at UW-Madison with funding from the Wisconsin Alumni Research Foundation.

\balance
\bibliographystyle{plain}
\bibliography{memory}

\end{document}